\documentclass[10pt, conference, a4paper]{IEEEtran}

\usepackage{amsmath,amssymb,amsfonts}
\usepackage{graphicx}
\usepackage{pifont}

\usepackage{amsthm}
\usepackage{xcolor}
\usepackage[figuresright]{rotating}

\usepackage{graphicx}
\graphicspath{{/}{fig/}}

\usepackage{array}
\usepackage{textcomp}
\usepackage{xcolor}
\usepackage{multirow}
\usepackage{booktabs}

\usepackage{mathtools}
\usepackage{breqn}
\usepackage{float}

\usepackage{pgfplots}
\pgfplotsset{compat=1.7}
\usepgfplotslibrary{groupplots}

\usepackage[style=base]{caption}
\usepackage{subcaption}
\usepackage{enumerate}

\usepackage{multirow,tabularx}
\usepackage{hyperref}
\usepackage{flushend}
\usepackage{algorithmic}
\usepackage[vlined, ruled, shortend]{algorithm2e}

\newlength\figureheight
\newlength\figurewidth
\SetAlCapNameFnt{\footnotesize}
\SetAlCapFnt{\footnotesize}

\usepackage[left=1.91cm,right=1.31cm,top=3.67cm,bottom=1.91cm]{geometry}

\title{
    \vspace{6.3mm}
    Hyperledger Fabric Blockchain and ROS\,2 Integration for Autonomous Mobile Robots \\
}

\author{
    \IEEEauthorblockN{
        Salma Salimi\IEEEauthorrefmark{2},
        Jorge Peña Queralta\IEEEauthorrefmark{2},
        Tomi Westerlund\IEEEauthorrefmark{2},
    }
    \IEEEauthorblockA{
        \normalsize
        \IEEEauthorrefmark{2}\href{https://tiers.utu.fi}{Turku Intelligent Embedded and Robotic Systems (TIERS) Lab, University of Turku, Finland}.\\
        Emails: \textsuperscript{1}\{salmas, jopequ, tovewe\}@utu.fi\\[-.42em]
    }
}

\begin{document}

\maketitle
\thispagestyle{empty}
\pagestyle{empty}



\begin{abstract}%
    \label{sec:abstract}
    In industrial applications, security and trust in the system are requirements for widespread adoption. Blockchain technologies have emerged as a potential solution to address identity management and secure data aggregation and control. However, the vast majority of works to date utilize Ethereum and smart contracts that are not scalable or well suited for industrial applications. This paper presents what is, to the best of our knowledge, the first integration of ROS~2 with the Hyperledger Fabric blockchain. With a framework that leverages Fabric smart contracts and ROS~2 through a Go application, we delve into the potential of using blockchain for controlling robots, and gathering and processing their data. We demonstrate the applicability of the proposed framework to an inventory management use-case where different robots are used to detect objects of interest in a given area. Designed to meet the requirements of distributed robotic systems, we show that the performance of the robots is not impacted significantly by the blockchain layer. At the same time, we provide examples for developing other applications that integrate Fabric smart contracts with ROS~2. Our results pave the way for further adoption of blockchain technologies in autonomous robotic systems for building trustable data sharing.
\end{abstract}

\IEEEpeerreviewmaketitle


\section{Introduction}\label{sec:introduction}

Autonomous robots are in turn more connected and distributed and has brought an increasing awareness in areas such as security of robotic systems~\cite{mayoral2022robot}
, the explainability and auditability of their behaviour~\cite{queralta2020enhancing}. At the same time, distributed ledger technologies (DLTs) are penetrating the design and development of networked systems and next-generation internet. At the intersection of these two trends, there is significant potential for the design of more secure and trustable distributed robotic systems, from swarms of robots~\cite{ferrer2021following} to autonomous vehicles~\cite{jain2021blockchain}. Surveillance or industrial application areas are an area of particular potential~\cite{de2021towards}.%

\begin{figure}
    \centering
    \includegraphics[width=0.49\textwidth]{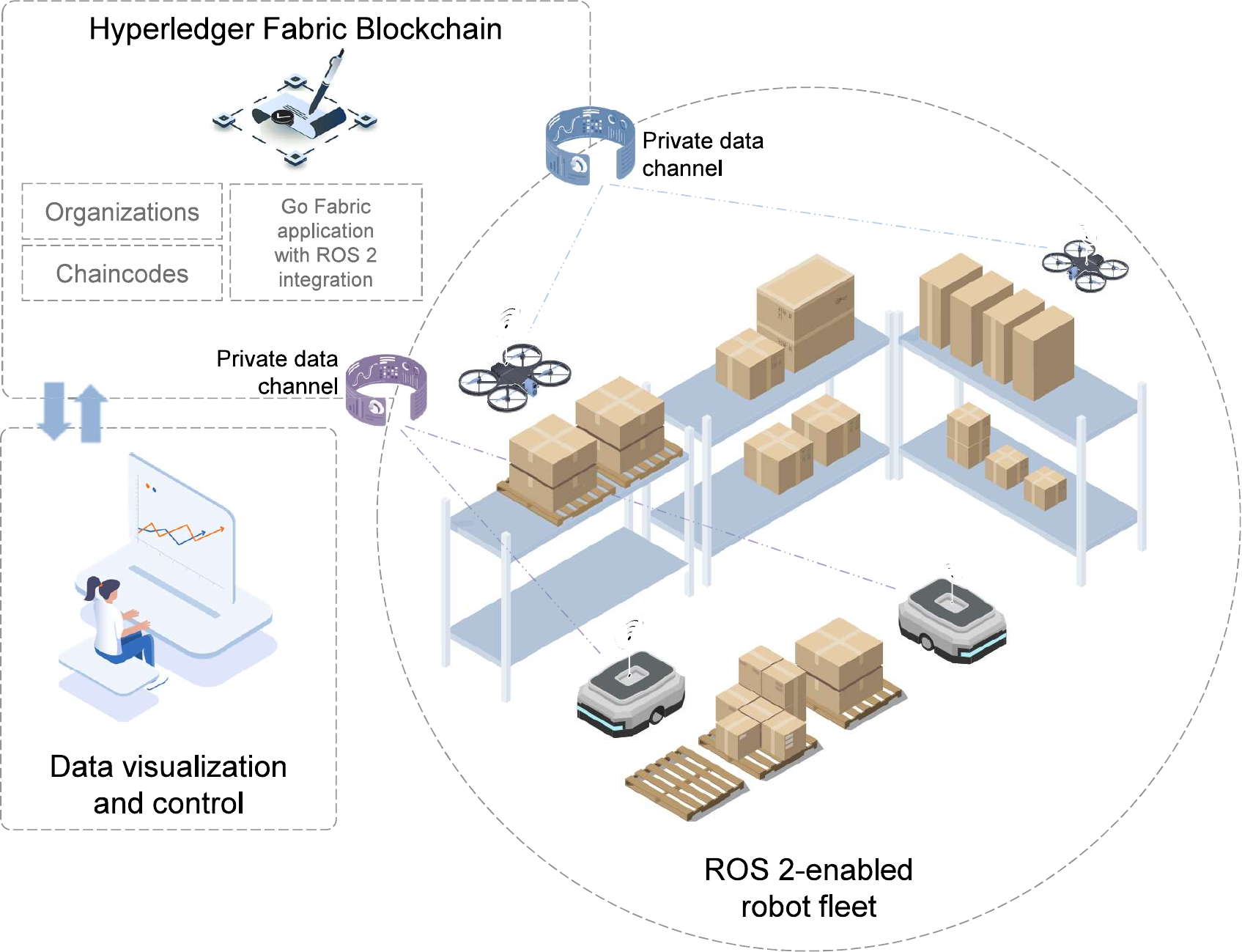}
    \caption{Conceptual view of the proposed application scenario, where a Hyperledger Fabric DLT network is interfaced with ROS~2 for managing autonomous robotic fleets.}
    \label{fig:concept}
\end{figure}

We are particularly interested in distributed robotic systems. The complexity of tasks and performance requirements have led to a growing interest in multi-robot systems (MRSs)~\cite{deng2021investigation}. The applications are multiple and in a variety of scenarios, e.g, emergency or surveillance~\cite{madridano2021trajectory}. Key aspects of such systems that can benefit from integrating DLTs are identity management, built-in security in data sharing, and multi-robot consensus. In multi-robot systems, knowledge-sharing-based collaborative inference need to often be conducted, so that complex tasks can be done. As untrusted data and security threaten the integrity of such data sharing, integrating DLTs provides a trustable solution that has proven effective at enhancing security while ensuring privacy and integrity~\cite{li2021blockchain}.%

Within the different DLT solutions that have been proposed in the literature, the most widely used so far is Ethereum~\cite{strobel2018managing, ferrer2021following}. Smart contracts running in the Ethereum Virtual Machine have enabled applications from secure federated learning~\cite{ferrer2018robochain} to trustable vehicular networks and other collaborative applications~\cite{xianjia2021flsurvey}. However, one of the key issues with Ethereum is its sustainability and scalability, with the standard consensus algorithms relying in cryptographic proof of work~\cite{queralta2021blockchain}. In addition, Ethereum is part of the family of public and permissionless blockchain solutions that do not necessarily fit the needs of industrial applications where more control over the data flows and robot identities is needed. Permissioned blockchain frameworks where only authorized nodes are allowed to share and access data are thus a good solution that maintains the immutability and security properties while ensuring more robust access control. Other alternatives for higher scalability can be found within next-generation blockchain solutions such as IOTA~\cite{de2021towards}. It is also worth noting the efforts being put in Ethereum 2.0 towards a more sustainable and scalable public blockchain~\cite{ethereum20specification}.

In this work, we propose the utilization of a permissioned, private blockchain for (i) managing the identity of robots in a fleet; (ii) interface user commands; (iii) store data samples for auditability of the system; (iv) share robot states securely within the fleet; and (v) control data access via smart contracts (see Fig.~\ref{fig:concept}). We utilize Hyperledger Fabric, an open-source blockchain solution that is highly modular and configurable, and specifically designed for industrial systems. It is designed as a foundation for developing applications and solutions with a modular architecture. Indeed, it allows components, such as consensus and membership services, to be plug-and-play. Also, it offers a unique approach to the consensus that enables performance at scale while preserving privacy~\cite{dalla2021your}.

In addition to the blockchain solution, we consider only integration with the latest version of the Robot Operating System (ROS~2), which supports natively distributed robotics systems, real-time control, and is ready for industrial applications. It is thus within our objectives in this work to provide an easy way to integrate a Fabric application with ROS~2.

The core contributions of this work are: (i) the introduction of a framework for integrating ROS2 with a Hyperledger Fabric Blockchain for distributed robotic systems; (ii) the analysis of the impact of integrating the Fabric blockchain in a robotic system together with a performance and scalability study; and (iii) an experimental proof of concept of the proposed framework for an inventory management application with ground and aerial robots.


\section{Related Works}
\label{sec:related_work}

The key properties from blockchain systems that can be leveraged in distributed multi-robot systems are built-in security aiding in data sharing within networked systems~\cite{sankar2017survey}, immutability of the data enabling auditability~\cite{white2019black}, and consensus protocols enabling collaborative decision-making through smart contracts~\cite{nguyen2019blockchain}.

An early integration of blockchain focused around the data immutability properties is the Black Block Recorder (BBR)~\cite{white2019black}. In this approach, distributed ledgers were leveraged for integrity proofs to enable tamper-evident logging, while considering the limited resources available for mobile robotic deployments. In more general terms, the potential of blockchains for achieving consensus in robot swarms and detecting potentially byzantine agents has been an active area of research~\cite{strobel2018managing, ferrer2021following}. In addition to these, other works have shown that parts of the blockchain stack such as Merkle trees can be leveraged for designing encoded sets of mission instructions for robotic systems~\cite{castello2019merkle}.

A variety of other studies show the wide range of potential application scenarios and integration possibilities. Guo et al. proposed a decentralized method for spherical amphibious multi-robot control systems based on blockchain technology~\cite{guo2021study}. A point-to-point (P2P) information network based on LORA technology has been created for this purpose, as well as embedded application environment and decentralized hardware and software architectures for multi-robot control systems based on blockchain technology. In heterogeneous multi-robot systems, optimizing the amount and type of data shared between robots with different sensing capabilities and computational resources is a key challenge, where smart contracts can be used for ranking data quality and managing computational and networking resources.~\cite{queralta2019blockchain}. A blockchain-based framework for collaborative edge knowledge inference for edge-assisted multi-robot systems was presented by Li et al.~\cite{li2021blockchain}. The authors proposed a knowledge-based blockchain consensus method for ensuring the trust of knowledge sharing and conducted a case study on emergency rescue applications.

In~\cite{lee2021upper}, Lee et al. presented a general solution based on a decentralized Monte Carlo tree search for scout-task coordination and an upper confidence bound for simultaneous exploration based on mutual information. The authors evaluated the performance of the algorithm in a multi-drone surveillance scenario in which scout robots use low-resolution, long-range sensors while task robots use short-range sensors to capture detailed information.

Moving the focus towards industrial application, Sah et al. proposed in~\cite{saha2021dhacs} a blockchain-based framework called Decentralized Hybrid Access Control for Smart contract (DHACS) for the Industrial Internet-of-Thing (IIoT) to bring up a robust access control mechanism for the IIoT. In another work, Lin et al. considered the challenge of secure data aggregation for the increasing number of data processing and sharing flows via industrial applications and services~\cite{lin2020blockchain}. The authors presented a blockchain-based privacy-aware distributed collection (BPDC) oriented strategy. With BPDC, they achieved privacy protection by decomposing sensitive tasks and task receivers into multiple groups, while guaranteeing the data aggregation performance. 


\section{Background}
\label{sec:background}

\subsection{Blockchain}

The most popular blockchains to date, Bitcoin and Ethereum, are permissionless Blockchain systems, where anonymous nodes can join and view all the data recorded on the blockchain without needing specific permission. As a result, the original blockchain technology cannot be directly applied to scenarios where particular organizations may want to join for the purposes of transacting with each other without exposing their data to the public or even to other parties sharing the same blockchain infrastructure. In permissioned architectures, such as Hyperledger, a subset of the peers is trusted and not all the nodes hold equal roles~\cite{androulaki2018hyperledger}. This allows for utilization of the blockchain technology stack in more controlled environments. Permissioned blockchain systems are able to protect data privacy confidentiality by putting in place multiple access control mechanisms and enabling only authorized participants to join the permissioned system and transact privately~\cite{wang2021private}. Most permissioned blockchains utilize deterministic consensus mechanisms, which can easily lead to fast consensus among authenticated users. These systems are therefore well suited to enterprise applications that require the processing of large volumes of transactions in a deterministic way~\cite{xu2021latency}.

\subsection{Hyperledger Fabric}

\begin{figure*}[t]
    \centering
    \includegraphics[width=.94\textwidth]{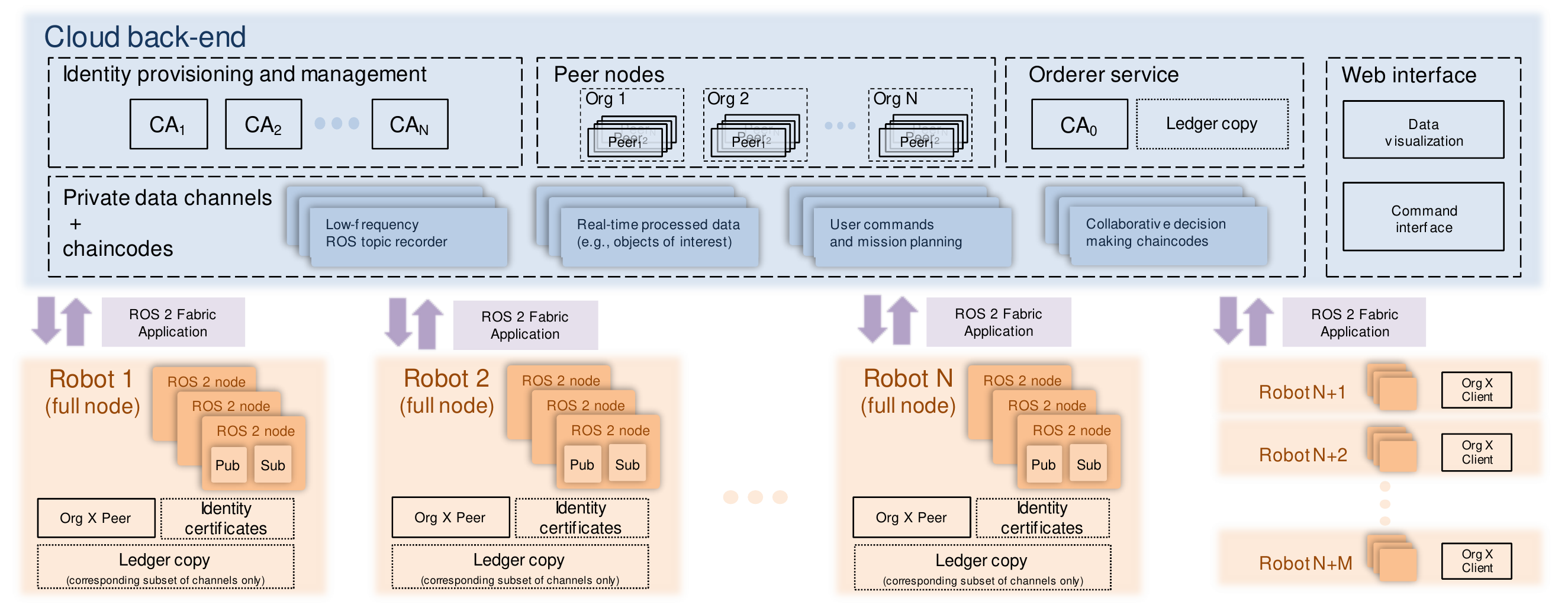}
    \caption{Architectural diagram of the proposed framework.}
    \label{fig:framework_architecture}
    \vspace{-.5em}
\end{figure*}

Hyperledger Fabric has been designed for enterprise use from the outset, with a set of characteristics that can be exploited in distributed robotic systems: (i) participants are identified, thus providing the tools for identity provisioning and management and certificate generation; (ii) networks are permissioned, meaning that a built-in layer of data security is readily available; (iii) high transaction throughput performance can meet the needs of real-time robotic data and data processing needs; (iv) with configurable low-latency transaction confirmation, consensus can be achieved in real-time with the networking capabilities as the limiting factor; and (v) data channel partitioning and privacy and confidentiality of transactions offer a seamless extension of ROS~2 topics, and the corresponding pub/sub system in the underlying DDS communication.


\section{ROS~2 + Fabric Framework}

\subsection{Framework architecture and key components}

The main components are illustrated in Fig.~\ref{fig:framework_architecture}. The core Fabric network is hosted in the cloud back-end, together with a Go-based web interface for visualizing ROS~2 data that has been saved in the different channels, as well as a command interface for inputting user instructions.

The Fabric network is hosted by a set of \textit{organizations}, each represented through a series of \textit{peer nodes}. In general terms, organizations are the containers for the peers and \textit{certificate authorities} (CA). Organizations have peers and CAs used to verify their membership in the network and are also called members of the network. The CAs are the certificate authorities through which every operation executed inside Hyperledger Fabric must be cryptographically signed. They generate the necessary certificates for the nodes, organizations definitions and applications of its organization. CAs play a key role in the network, because they are trusted to identify components as belonging to a specific organization.

One of the key components of Fabric networks that differentiate it from other blockchain solutions is the existence of private data \textit{channels}. These allow for the network to be partitioned while maintaining a global ledger state. Within channels, \textit{chaincodes} are deployed for supporting \textit{smart contracts}. A chaincode definition is used by organizations to agree on the parameters of the chaincode before it can be used on a channel. Each channel member that wants to use the chaincode then needs to approve its definition for their organization. Once enough channel members have approved a chaincode definition, it can be committed to the channel. After the definition is committed, the first invokation of the chaincode will commit its state on the corresponding channel.

Finally, smart contracts are a common set of functions which must be drafted before peers from different organizations can transact with each other. They contain agreements cover common terms, data, rules, concept definitions, and processes. They are often invoked by an application external to the blockchain and provide an interface for interacting with the ledger. Both the applications and smart contracts can be written in general-purpose programming languages such as Go, Java or Node.js~\cite{polge2021permissioned}.

\subsection{Integration of Fabric applications and ROS~2 nodes}

As illustrated in Fig.~\ref{fig:framework_architecture}, robots in the proposed framework are both members of the Fabric network and a potentially shared ROS~2 network. In the figure, we have listed the most significant applications of smart contracts for industrial robot fleets. First, as identified already in the literature~\cite{white2019black}, the ledger can be used to store immutable records of data, either sample sensor data or other standard ROS data types, for example. An example of an application for such a purpose is illustrated in Algorithm~\ref{alg:ros2_fabric_recorder} Real-time processed data can also be stored through smart contracts to trigger predefined actions at other modules. Finally, any collaborative decision making process can be implemented through smart contracts to ensure that all robots obtain the same result. This is applicable, e.g., to role allocation or resource distribution problems.

The rest of the framework includes a web application for visualizing data and sending commands to the mobile robots. Specifically, the list of assets in a given channel can be browsed, and position data from the robots visualized in a map together with locations of detected objects of interest. Sensor data samples can also be directly viewed such as images.

\begin{algorithm}[t]
    \small
	\caption{Low-frequency ROS data recorder}
	\label{alg:ros2_fabric_recorder}
	\KwIn{\\
	    Topic name: \textit{data\_topic}\\
	    Recording frequency: \textit{max\_freq}\\
	}
	\textbf{Initialization:}\\
	\hspace{1em}request\_bring\_up\_network();\\
	\hspace{1em}request\_create\_channel();\\
	\hspace{1em}request\_deploy\_chaincode\_to\_channel();\\
    \While {network is up} {
        \If{chaincode deployed to the channel} {
            Assert smart contract's API is accessible;\\
            Initialize ROS2 node;\\
            \ForEach{recording application} {
                $load\_wallet\_and\_identity()$;\\
                $connect\_to\_gateway()$;\\
                $connect\_to\_network()$;\\
                $connect\_to\_channel()$;\\
                $load\_chaincode()$;\\
                $subscribe\_to\_ROS2\_topic$(\\
                \hspace{1em}callback(msg) = func(\{\\
                    \hspace{2em}\textbf{if}\,{ros\_time - last\_msg\_time $\geq$ 1/max\_freq}\\
                    \hspace{2em}\textbf{then}\\
                        \hspace{3em}data $\leftarrow$ deserialize\_msg();\\
                        \hspace{3em}create\_chaincode\_asset(data);\\
                        \hspace{3em}\footnotesize{\tcp{Optionally download data:}}
                        \hspace{3em}recorder\_data $\leftarrow$ download\_channel\_assets();\\
                \hspace{1em}\})\\
                );\\
            }
        }
    }
\end{algorithm}
\begin{figure*}
    \centering
    \includegraphics[width=.92\textwidth]{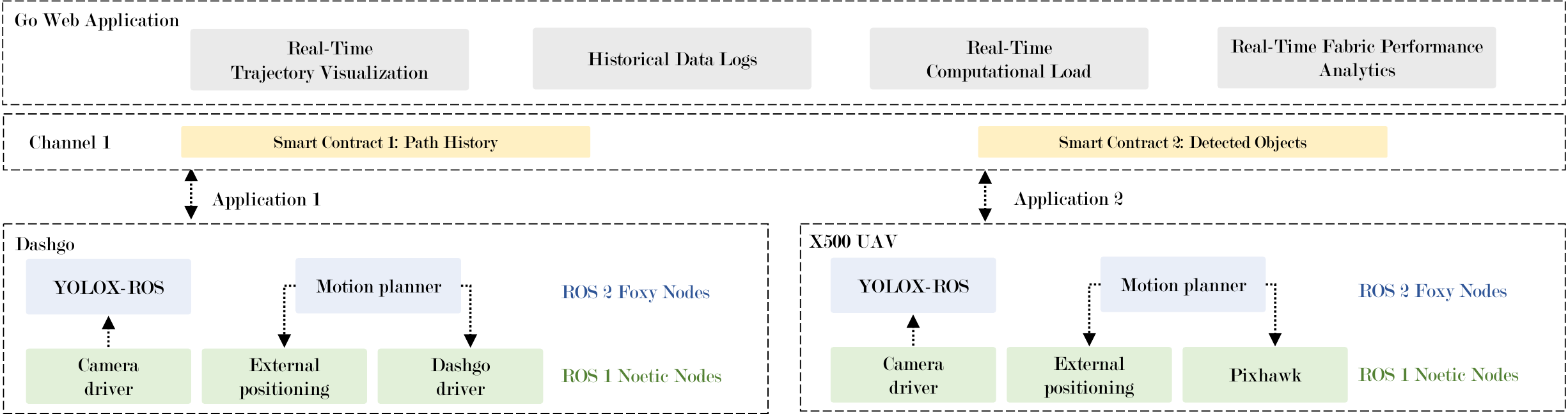}
    \caption{{Implementation Diagram}}
    \label{fig:implementation}
\end{figure*}

\section{Experimental Results}

For the rest of the paper, we focus on a proof of concept of the proposed framework with an inventory management use case. We demonstrate the usability of the proposed framework with an experiment where a set of ground and aerial robots are used for object detection in a warehouse-like environment. Hyperledger Fabric is used as the backbone for storing data and controlling the robots through a Go application and integration of smart contracts with ROS2. A web interface enables real-time tracking of the robot's trajectory and detected objects.

\subsection{Experimental setup}

\subsubsection{Heterogeneous Multi-Robot System} 

the employed multi-robot system in this paper consists of a ground robot and a unmanned aerial vehicle (UAV). The ground robot is an EAI Dashgo platform equipped with an UP HD camera with an OV2735 sensor.

The custom-built UAV is based on the X500 quad-rotor frame which is embedded with a Pixhawk 5X flight controller running the PX4 firmware. A TF Mini Lidar is utilized for height estimation on the UAV, also equipped with an UP HD camera with an OV2735 sensor. An AAEON Up Xtreme with an Intel i7-8665UE processor is used as a companion computer on both the Dashgo and the UAV. Both robots use RealSense T265 cameras for VIO-based egomotion estimation. For this experiment we have relied on an external motion capture system with six Optitrack PrimeX 22 cameras for navigation.

\subsubsection{Software} 

robots are running ROS Noetic under Ubuntu 20.04 for the main drivers. Localization and object detection are running in ROS~2 Foxy. A diagram of the different software modules running in different nodes is shown in Fig.~\ref{fig:implementation}. The fabric applications runs onboard the drones but connects to peers running on a separate computer in the network with the same Intel i7 processor.

The Dashgo platform is controlled with the manufacturer's driver, while MAVROs is run in ROS Noetic for the control of the UAV in offboard mode. Data from the optitrack system is received with a VRPN client ROS node and forwarded to MAVROS for waypoint control. A simple motion planner for the Dashgo has been written for this experiment.

The $ros1\_bridge$ package is used to forward data from Noetic to Foxy topics under the same computer. The $usb\_cam$ package available in both Noetic and Foxy is used to obtain camera images at a frequency of 30\,Hz, even though they are forwarded to the object detector at 5\,Hz only as this is sufficient for the proof of concept. The object detector used in the experiments is YOLOX, and a selection of objects part of the categories in the COCO dataset is used for the purpose of the inventory management.

\subsubsection{Fabric channels and smart contracts}
for the implementation of the different parts of the system, consisting the smart contracts, the application to interact with them, the ROS~2 nodes and the web interface, the Go programming language (golang) has been used.This helps in the integration process as a single Go module can be connected to both ROS~2 through \textit{rclgo}, the ROS~2 GO client library. Sample codes for all these components are made available in the project's repository\footnote{https://github.com/TIERS/ros2-fabric-integration}. Also a private Hyperledger Fabric network has been brought up for secure data management and robot control.

Two smart contracts, one for storing the path tracking of the robots and one for storing the location of the detected objects in the asset have been used in this implementation (see Fig.~\ref{fig:implementation}). For each robot, one application has been used which has various functions for controlling the assets, such as creating new assets, read all the assets, check existence of the assets, updating the assets and also changing them.

\subsection{Scalability and performance results}

The two robots are commanded to follow predefined paths and store in a Fabric channel chaincode a series of objects for the purpose of managing the inventory. For this, a series of shelves are set up in a room of $40\,m^2$ with various objects from the COCO dataset categories.

\begin{figure}
    \centering
    \setlength\figureheight{0.30\textwidth}
    \setlength\figurewidth{0.45\textwidth}
    \scriptsize{\input{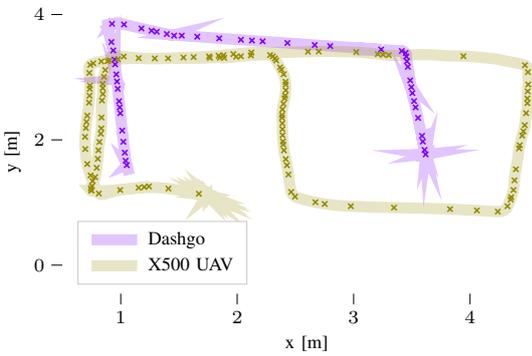}}
    \caption{Trajectory tracking of both robots and detected objects where the location of the detections are marked in darker color.}
    \label{fig:drone_trajectory}
\end{figure}

Figure~\ref{fig:drone_trajectory} shows the trajectories of the robots during the experiment, together with the set of locations where objects of interest were detected. For each of these, a new asset was generated in the corresponding channel. In addition to that, the robot's trajectory is sampled at $0.2\,Hz$.

\begin{figure}
    \centering
    \setlength\figureheight{0.23\textwidth}
    \setlength\figurewidth{0.45\textwidth}
    \scriptsize{\input{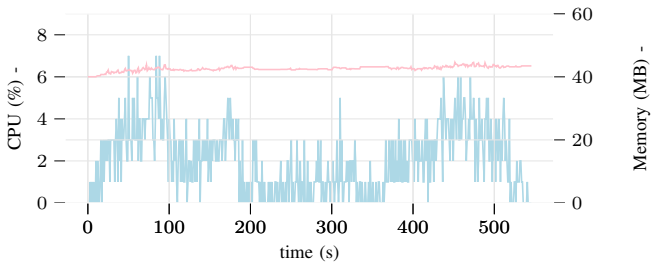}}
    \caption{Go application activity during the mission where CPU usage is shown in blue and memory in pink.}
    \label{fig:app_activity}
\end{figure}

\begin{figure}
    \centering
    \setlength\figureheight{0.23\textwidth}
    \setlength\figurewidth{0.44\textwidth}
    \scriptsize{\input{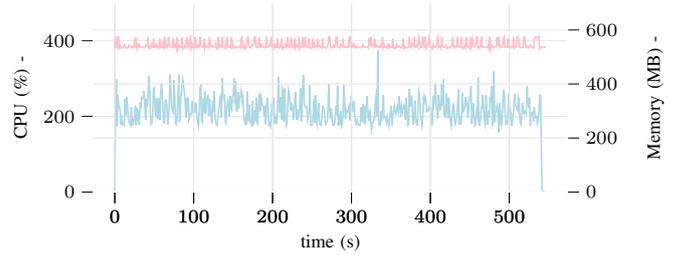}}
    \caption{YOLOX ROS~2 node activity during the mission where CPU usage is shown in blue and memory in pink.}
    \label{fig:yolox_activity}
\end{figure}

The system CPU and memory usage during the experiments is shown in Fig.~\ref{fig:app_activity}, and the corresponding YOLOX resource utilization is shon in Fig.~\ref{fig:yolox_activity}. The objective of this analysis is to quantify the impact, in terms of computational resources, that integrating the Fabric network might have into an existing ROS~2 system. From these results, we can conclude that the addition of Fabric as an additional channel for data sharing is negligible, and therefore the proposed framework has potential to be adopted in a wide variety of application scenarios and robotic platforms.


Finally, in order to assess the scalability and performance of the system under a more realistic workload, we have performed a series of stress tests. In these tests, data was transmitted from ROS~2 to the Fabric blockchain at very high frequencies, in order to calculate the latency that high loads induce in the system, as well as the ability of the Fabric network to process such high volumes. We show in Table~\ref{tab:performance} the results of a subset of 8 of the stress tests with different Fabric configurations. These have been selected as the most representative from a wider set of over 20 tests. The maximum transaction throughput that we obtain is close 200\,Hz with the standard network configuration and modifying only the batch timeout and maximum messages parameters. These define either the maximum timeout after a transaction has happened before a block is generated, or the maximum number of messages that are accumulated before mining a block. Transactions (e.g., assets being created) are only confirmed once they appear in a block, and therefore this becomes a key parameter affecting the system latency. We can in general observe from the table that Fabric is most optimal with small blocks, something that has already been identified in the literature~\cite{xu2021latency}. The use of computational resources (CPU, RAM) for the stress tests in the most representative cases is shown in Fig.~\ref{fig:ram_cpu}.

In addition to the resource utilization and transaction throughput, we have measured the actual latency of storing data in the blockchain when the client application node was running on the robots, connected through Wi-Fi to the computer running the peer and orderer nodes. The distribution of the latency of the main four settings is reported in Fig.~\ref{fig:latency}, where data from 15\,s is accumulated and over 200\,Hz of ROS~2 data being pushed into the smart contract.

\begin{table}[t]
    \centering
    \caption{Results of stress tests under different fabric configurations to assess the network's transaction throughput limitations.}
    \label{tab:performance}
    \footnotesize
    \begin{tabular}{@{}lcccc@{}}
        \toprule
        Fabric & Batch & Max & Avg. transaction & Avg. orderer \\
        configuration & Timeout & Messages & throughput & CPU load \\
        \midrule
        1 & 5 & 5 & 183\,Hz & 28\,\% \\
        2 & 5 & 10 & 118\,Hz & 28\,\% \\
        3 & 5 & 20 & 69\,Hz & 28\,\% \\
        4 & 5 & 100 & 29\,Hz & 23\,\% \\
        5 & 5 & 1000 & 28\,Hz & 23\,\% \\
        6 & 0.1 & 100 & 70\,Hz & 28\,\% \\
        7 & 0.05 & 100 & 114\,Hz & 28\,\% \\
        8 & 0.025 & 100 & 155\,Hz & 28\,\% \\
        \bottomrule
    \end{tabular}
\end{table}
\begin{figure}
    \centering
    \begin{subfigure}{.48\textwidth}
        \setlength\figureheight{0.6\textwidth}
        \setlength\figurewidth{\textwidth}
        \scriptsize{
\begin{tikzpicture}

\definecolor{chocolate2267451}{RGB}{226,74,51}
\definecolor{dimgray85}{RGB}{85,85,85}
\definecolor{gainsboro229}{RGB}{229,229,229}
\definecolor{gray}{RGB}{128,128,128}
\definecolor{gray119}{RGB}{119,119,119}
\definecolor{lightgray204}{RGB}{204,204,204}
\definecolor{mediumpurple152142213}{RGB}{152,142,213}
\definecolor{sandybrown25119394}{RGB}{251,193,94}
\definecolor{steelblue52138189}{RGB}{52,138,189}

\begin{axis}[
    height=\figureheight,
    width=\figurewidth,
    axis background/.style={fill=white!92!black},
    axis line style={white},
    legend cell align={left},
    legend style={
      fill opacity=0.8,
      draw opacity=0.666,
      text opacity=1,
      draw=white!80!black
    },
    legend pos=south east,
    tick align=outside,
    tick pos=left,
    x grid style={white},
    xmajorgrids,
    xmin=0, xmax=15,
    xtick style={color=white},
    xticklabels={,},
    y grid style={white},
    ylabel=\textcolor{dimgray85}{CPU (\%) -},
    ymajorgrids,
    ymin=-5, ymax=103,
    ytick style={color=black}
]
\addplot [semithick, steelblue52138189]
table {%
0 0
1 95
2.001 93.9
3.002 94.9
4.004 94.9
5.005 95.9
6.005 96
7.007 95.9
8.008 96.9
9.009 94.9
10.01 95
11.011 93.9
12.012 95.9
13.013 96.9
14.013 94.9
15.014 94.9
16.016 93.9
17.017 96.9
};
\addlegendentry{Fabric BT0.025_M100}
\addplot [semithick, mediumpurple152142213]
table {%
0 0
1.001 90.9
2.003 87.9
3.003 88
4.004 87
5.004 86
6.005 85.9
7.006 84.9
8.008 85.9
9.009 84.9
10.01 84.9
11.01 86
12.012 84.9
13.012 84.9
14.014 86.9
15.015 85.9
16.015 88.9
17.016 84
18.016 86
19.017 85.9
20.018 84.9
21.018 84.9
22.019 83.9
23.021 84.9
24.022 83.9
25.023 87.9
26.025 83.9
27.025 83
28.026 81.9
29.026 85.9
};
\addlegendentry{Fabric BT5_M5}
\addplot [semithick, gray119]
table {%
0 0
1 86
2.001 86
3.001 88
4.003 85.9
5.003 89.9
6.004 81.9
7.005 86
8.006 86.9
9.007 86.9
10.008 86
11.009 87.9
12.011 85.9
13.012 85.9
14.013 85.9
15.014 85.9
16.015 85.9
17.016 84.9
18.017 85.9
};
\addlegendentry{Fabric BT5_M100}
\addplot [semithick, chocolate2267451]
table {%
0 0
1 88
2.001 84.9
3.003 89.9
4.003 86.9
5.004 85.9
6.005 87.9
7.005 85
8.007 88.9
9.008 87.9
10.009 85.9
11.009 89
12.011 84.9
13.012 86.9
14.013 86.9
15.014 88
16.015 84.9
17.016 84.9
18.017 87.9
19.019 85.9
20.02 85.9
21.021 87.9
22.022 87
23.022 89
24.023 85.9
};
\addlegendentry{Fabric BT5_M1000}
\end{axis}

\end{tikzpicture}}
    \end{subfigure}
    
    \vspace{-1em}
    \hspace{-1.3em}
    \begin{subfigure}{.48\textwidth}
        \centering
        \setlength\figureheight{0.42\textwidth}
        \setlength\figurewidth{\textwidth}
        \scriptsize{
\begin{tikzpicture}

\definecolor{chocolate2267451}{RGB}{226,74,51}
\definecolor{dimgray85}{RGB}{85,85,85}
\definecolor{gainsboro229}{RGB}{229,229,229}
\definecolor{gray}{RGB}{128,128,128}
\definecolor{gray119}{RGB}{119,119,119}
\definecolor{lightgray204}{RGB}{204,204,204}
\definecolor{mediumpurple152142213}{RGB}{152,142,213}
\definecolor{sandybrown25119394}{RGB}{251,193,94}
\definecolor{steelblue52138189}{RGB}{52,138,189}

\begin{axis}[
    height=\figureheight,
    width=\figurewidth,
    axis background/.style={fill=white!92!black},
    axis line style={white},
    legend cell align={left},
    legend style={fill opacity=0.8, draw opacity=1, text opacity=1, draw=white!80!black},
    tick align=outside,
    tick pos=bottom,
    x grid style={white},
    xlabel=\textcolor{dimgray85}{time (s)},
    xmajorgrids,
    xmin=0, xmax=15,
    xtick style={color=dimgray85},
    y grid style={white},
    ylabel=\textcolor{dimgray85}{Memory (MB) -},
    ymajorgrids,
    ymin=-5, ymax=103,
    ytick style={color=black} 
]
\addplot [semithick, steelblue52138189]
table {%
0 44.84
1 46.551
2.001 46.336
3.002 46.141
4.004 45.781
5.005 45.547
6.005 45.766
7.007 46.523
8.008 45.906
9.009 48.094
10.01 46.516
11.011 46.102
12.012 45.797
13.013 47.5
14.013 46.156
15.014 46.242
16.016 47.078
17.017 46.043
};
\addplot [semithick, mediumpurple152142213]
table {%
0 57.648
1.001 58.082
2.003 58.824
3.003 59.113
4.004 59.461
5.004 60.824
6.005 60.984
7.006 61.344
8.008 63.164
9.009 63.031
10.01 63.441
11.01 63.965
12.012 64.75
13.012 65.652
14.014 66.434
15.015 68.332
16.015 68.16
17.016 69.238
18.016 69.336
19.017 69.809
20.018 70.117
21.018 70.797
22.019 71.047
23.021 72.285
24.022 72.844
25.023 73.09
26.025 74.578
27.025 74.234
28.026 74.234
29.026 74.488
};
\addplot [semithick, gray119]
table {%
0 46.797
1 48.809
2.001 47.375
3.001 48.594
4.003 47.355
5.003 48.496
6.004 47.406
7.005 47.395
8.006 47.488
9.007 47.344
10.008 47.488
11.009 47.227
12.011 49.668
13.012 47.02
14.013 46.875
15.014 46.867
16.015 46.844
17.016 47.398
18.017 48.688
};
\addplot [semithick, chocolate2267451]
table {%
0 47.422
1 46.383
2.001 46.684
3.003 47.062
4.003 47.141
5.004 46.836
6.005 47.008
7.005 47.301
8.007 46.57
9.008 47.754
10.009 46.855
11.009 49.172
12.011 47.641
13.012 47.848
14.013 49.691
15.014 47.531
16.015 50.137
17.016 47.664
18.017 46.688
19.019 46.59
20.02 46.867
21.021 48.344
22.022 47.473
23.022 46.781
24.023 48.656
};
\end{axis}

\end{tikzpicture}}
    \end{subfigure}
    \caption{Memory and CPU usage of the computer running the orderer and peer containers during the stress tests, where BT is the batch timeout (in seconds) and M is the max. messages to add in a block.}
    \label{fig:ram_cpu}
    \vspace{-.42em}
\end{figure}
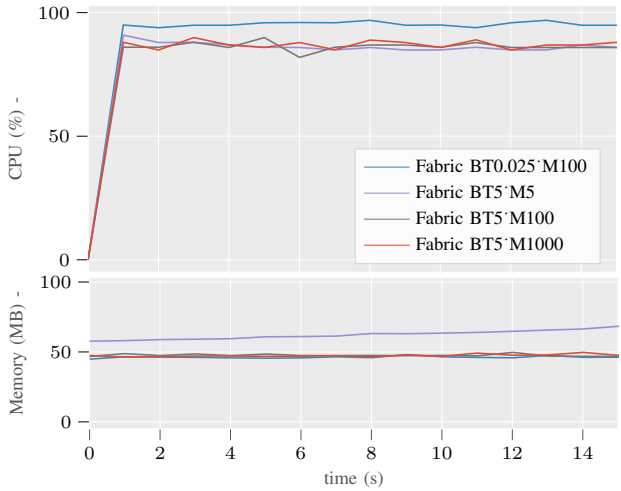
\begin{figure}[t]
    \centering
    \setlength\figureheight{0.23\textwidth}
    \setlength\figurewidth{0.49\textwidth}
    \scriptsize{
\begin{tikzpicture}

\definecolor{darkgray176}{RGB}{176,176,176}
\definecolor{darkorange25512714}{RGB}{255,127,14}
\definecolor{steelblue31119180}{RGB}{31,119,180}

\definecolor{color0}{rgb}{1,0.752941176470588,0.796078431372549}
\definecolor{color1}{rgb}{0.67843137254902,0.847058823529412,0.901960784313726}
\definecolor{color2}{rgb}{1,0.752941176470588,0.796078431372549}
\definecolor{color4}{rgb}{0.67843137254902,0.847058823529412,0.901960784313726}

\begin{axis}[
height=\figureheight,
width=\figurewidth,
axis line style={white},
legend style={fill opacity=0.8, draw opacity=1, text opacity=1, 
draw=white!80!black},
tick align=outside,
tick pos=left,
x grid style={white!69.0196078431373!black},
xmin=0.5, xmax=4.5,
xtick style={color=black},
xtick={1,2,3,4},
xticklabels={BT5_M5,BT5_M10,BT5_M100,BT5_M1000},
y grid style={white!90!black},
ymajorgrids,
ylabel={time (ms)},
ymin=-198.69630346035, ymax=4194.69728559335,
ytick style={color=black}
]
\addplot [black]
table {%
1 801.237331
1 727.277333
};
\addplot [black]
table {%
1 906.994927
1 971.761309
};
\addplot [black]
table {%
0.8875 727.277333
1.1125 727.277333
};
\addplot [black]
table {%
0.8875 971.761309
1.1125 971.761309
};
\addplot [black, mark=+, mark size=3, mark options={solid}, only marks]
table {%
1 326.966269
1 323.962805
1 318.815322
1 322.386241
1 317.37872
1 573.695077
1 568.771473
1 563.699699
1 558.689688
1 553.608896
1 548.545784
1 543.450123
1 538.368379
1 533.305753
1 528.257651
1 523.198498
};
\addplot [black]
table {%
2 580.944449
2 535.920306
};
\addplot [black]
table {%
2 621.32922775
2 681.242792
};
\addplot [black]
table {%
1.8875 535.920306
2.1125 535.920306
};
\addplot [black]
table {%
1.8875 681.242792
2.1125 681.242792
};
\addplot [black, mark=+, mark size=3, mark options={solid}, only marks]
table {%
2 690.610412
2 685.656701
2 725.590955
2 720.625544
2 715.656983
2 713.444541
2 708.360772
2 703.335807
2 700.163934
2 695.223661
2 690.228076
2 685.115482
};
\addplot [black]
table {%
3 1680.63340025
3 1.003405133
};
\addplot [black]
table {%
3 3442.09834575
3 3893.07945
};
\addplot [black]
table {%
2.8875 1.003405133
3.1125 1.003405133
};
\addplot [black]
table {%
2.8875 3893.07945
3.1125 3893.07945
};
\addplot [black]
table {%
4 2085.128487
4 1023.290278
};
\addplot [black]
table {%
4 3362.672452
4 3994.997577
};
\addplot [black]
table {%
3.8875 1023.290278
4.1125 1023.290278
};
\addplot [black]
table {%
3.8875 3994.997577
4.1125 3994.997577
};
\path [draw=black, fill=color0, opacity=0.42]
(axis cs:0.775,801.237331)
--(axis cs:1.225,801.237331)
--(axis cs:1.225,906.994927)
--(axis cs:0.775,906.994927)
--(axis cs:0.775,801.237331)
--cycle;
\path [draw=black, fill=color1, opacity=0.42]
(axis cs:1.775,580.944449)
--(axis cs:2.225,580.944449)
--(axis cs:2.225,621.32922775)
--(axis cs:1.775,621.32922775)
--(axis cs:1.775,580.944449)
--cycle;
\path [draw=black, fill=color2]
(axis cs:2.775,1680.63340025)
--(axis cs:3.225,1680.63340025)
--(axis cs:3.225,3442.09834575)
--(axis cs:2.775,3442.09834575)
--(axis cs:2.775,1680.63340025)
--cycle;
\path [draw=black, fill=color4]
(axis cs:3.775,2085.128487)
--(axis cs:4.225,2085.128487)
--(axis cs:4.225,3362.672452)
--(axis cs:3.775,3362.672452)
--(axis cs:3.775,2085.128487)
--cycle;
\addplot [black]
table {%
0.775 840.625244
1.225 840.625244
};
\addplot [black]
table {%
1.775 600.1074385
2.225 600.1074385
};
\addplot [black]
table {%
2.775 2800.2474355
3.225 2800.2474355
};
\addplot [black]
table {%
3.775 2541.796267
4.225 2541.796267
};
\end{axis}

\end{tikzpicture}}
    \caption{Distribution of the latency for committing transactions between the robot and the peer node connected through Wi-Fi.} 
    \label{fig:latency}
\end{figure}
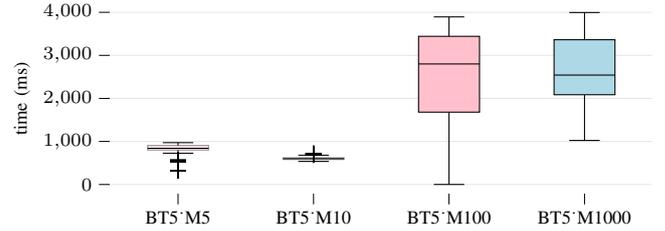


\section{Conclusion}
\label{sec:conclusion}

In this work, we present a framework for integrating ROS~2 with a Hyperledger Fabric blockchain for the purposes of identity management, secure control interfaces, auditable data flows and private data channels in industrial robot fleets. In comparison with the literature in the integration of blockchain in robotic systems, we expect this work to fill the gap in the use of permissioned blockchains. At the same time, we provide samples of applications that relay data between ROS~2 nodes and smart contracts in Fabric. Our results show that this solutions meets the needs of real-time distributed system, and provides a significant amount of built-in security and identity management features, while having a minimal impact on the utilization of computational resources. The system is demonstrated with a proof of concept through an inventory management use case with different types of mobile robots.



\bibliographystyle{unsrt}
\bibliography{new_bib}

\end{document}